# NOTES ON LR PARSER DESIGN

## Christer Samuelsson

Swedish Institute of Computer Science,
Box 1263 S–164 28 Kista, Sweden. E-mail: christer@sics.se

## 1 INTRODUCTION

This paper discusses the design of an LR parser for a specific high-coverage English grammar. The design principles, though, are applicable to a large class of unification-based grammars where the constraints are realized as Prolog terms and applied monotonically through instantiation, where there is no right movement, and where left movement is handled by gap threading.

The LR parser was constructed for experiments on probabilistic parsing and speedup learning, see [10]. LR parsers are suitable for probabilistic parsing since they contain a representation of the current parsing state, namely the stack and the input string, and since the actions of the parsing tables are easily attributed probabilities conditional on this parsing state. LR parsers are suitable for the speedup learning application since the learned grammar is much larger than the original grammar, and the prefixes of the learned rules overlap to a very high degree, circumstances that are far from ideal for the system's original parser. Even though these ends influenced the design of the parser, this article does not focus on these applications but rather on the design and testing of the parser itself.

## 2 LR PARSING

An LR parser is a type of shift-reduce parser originally devised by Knuth for programming languages [4]. The success of LR parsing lies in handling a number of grammar rules simultaneously, rather than attempting one at a time, by the use of prefix merging. LR parsing in general is well described in [1], and its application to natural-language processing in [12].

An LR parser is basically a pushdown automaton, i.e. it has a pushdown stack in addition to a finite set of internal states, and a reader head for scanning the input string from left to right, one symbol at a time. In fact, the "L" in "LR" stands for left-to-right scanning of the input string. The "R" stands for constructing the rightmost derivation in reverse.

The stack is used in a characteristic way: The items on the stack consist of alternating grammar symbols and states. The current state is the state on top of the stack. The most distinguishing feature of an LR parser is however the form of the transition relation — the action and goto tables. A non-deterministic LR parser can in each step perform one of four basic actions. In state `S` with lookahead symbol `Sym` it can:

1. `accept(S,Sym)`: Halt and signal success.
2. `shift(S,Sym,S2)`: Consume the symbol `Sym`, place it on the stack, and transit to state `S2`.
3. `reduce(S,Sym,R)`: Pop off a number of items from the stack corresponding to the RHS of grammar rule `R`, inspect the stack for the old state `S1`, place the LHS of rule `R` on the stack, and transit to state `S2` determined by `goto(S1,LHS,S2)`.
4. `error(S,Sym)`: Fail and backtrack.

Prefix merging is accomplished by each internal state corresponding to a set of partially processed grammar rules, so-called "dotted items" containing a dot ($\cdot$) to mark the current position. Since the grammar of Fig. 1 contains Rules 2, 3, and 4, there will be a state containing the dotted items

$$VP \rightarrow V \cdot$$
$$VP \rightarrow V \cdot NP$$
$$VP \rightarrow V \cdot NP\ NP$$

This state corresponds to just having found a verb ($V$). Which of the three rules to apply in the end will be determined by the rest of the input string; at this point no commitment has been made to either.

Compiling LR parsing tables consists of constructing the internal states (i.e. sets of dotted items) and from these deriving the shift, reduce, accept and goto entries of the transition relation. New states can be induced from previous ones; given a state `S1`, another state `S2` reachable from it by `goto(S1,Sym,S2)` (or `shift(S1,Sym,S2)` if `Sym` is a terminal symbol) can be constructed as follows:

1. Select all items in state `S1` where a particular symbol `Sym` follows immediately after the dot and move the dot to after this symbol. This yields the kernel items of state `S2`.

2. Construct the non-kernel closure by repeatedly adding a so-called non-kernel item (with the dot at the beginning of the RHS) for each grammar rule whose LHS matches a symbol following the dot of some item in `S2`.

Consider for example the grammar of Fig. 1, which will generate the states of Fig. 2. State 1 can be constructed from State 0 by advancing the dot in $S \rightarrow \cdot NP\ VP$ and $NP \rightarrow \cdot NP\ PP$ to form the items $S \rightarrow NP \cdot VP$ and $NP \rightarrow NP \cdot PP$, which constitute the kernel of State 1. The non-kernel items are generated by the grammar

$$
\begin{array}{rcll}
S & \rightarrow & NP\ VP & (1) \\
VP & \rightarrow & V & (2) \\
VP & \rightarrow & V\ NP & (3) \\
VP & \rightarrow & V\ NP\ NP & (4) \\
VP & \rightarrow & VP\ PP & (5) \\
NP & \rightarrow & Det\ N & (6) \\
NP & \rightarrow & Pron & (7) \\
NP & \rightarrow & NP\ PP & (8) \\
PP & \rightarrow & Prep\ NP & (9) \\
\end{array}
$$

Figure 1: A toy grammar

rules for VPs and PPs, the categories following the dot in the new items, namely Rules 2, 3, 4, 5 and 9.

Using this method, the set of all parsing states can be induced from an initial state whose single kernel item has the top symbol of the grammar preceded by the dot as its RHS (the item $S' \rightarrow \cdot S$ of State 0 in Fig. 2). The accept, shift and goto entries fall out automatically from this procedure. Any dotted item where the dot is at the end of the RHS gives rise to a reduction by the corresponding grammar rule. Thus it remains to determine the lookahead symbols of the reduce entries.

In Simple LR (SLR) the lookahead is any terminal symbol that can immediately follow any symbol of the same type as the LHS of the rule. In LookAhead LR (LALR) it is any terminal symbol that can immediately follow the LHS given that it was constructed using this rule in this state. In general, LALR gives considerably fewer reduce entries than SLR, and thus results in faster parsing. In the experiments this reduced the parsing times by 30 %.

## 3 PROBLEMS WITH LR PARSING

The problems of applying the LR-parsing scheme to large unification grammars for natural language, rather than small context-free grammars for programming languages, stem from three sources. The first is that symbol matching no longer consists of checking atomic symbols for equality, but rather comparing complex feature structures. The second is the high level of ambiguity of natural language and the resulting non-determinism. The third is the sheer size of the grammars.

Straight-forward resorting to a context-free backbone grammar and subsequent filtering using the full constraints of the underlying unification grammar (UG) is an approach taken by for example [3]. The problem with this approach is that the predictive power of the unification grammar is so vastly diluted when feature propagation is omitted. Firstly, the context-free backbone grammar will in general allow very many more analyses than the unification grammar, leading to poor parser performance. Secondly, the feature propagation necessary for gap threading to prevent non-termination due to empty productions is obstructed.

On the other hand, the treatment of the full UG constraints in the parsing-table construction phase is associated with a number of problems most of which

*State 0*
$S' \rightarrow \cdot S$
$S \rightarrow \cdot NP\ VP$
$NP \rightarrow \cdot Det\ N$
$NP \rightarrow \cdot Pron$
$NP \rightarrow \cdot NP\ PP$

*State 2*
$NP \rightarrow Det \cdot N$

*State 3*
$NP \rightarrow Pron \cdot$

*State 5*
$S \rightarrow NP\ VP \cdot$
$VP \rightarrow VP \cdot PP$
$PP \rightarrow \cdot Prep\ NP$

*State 7*
$NP \rightarrow NP\ PP \cdot$

*State 8*
$PP \rightarrow Prep \cdot NP$
$NP \rightarrow \cdot Det\ N$
$NP \rightarrow \cdot Pron$
$NP \rightarrow \cdot NP\ PP$

*State 11*
$VP \rightarrow V\ NP \cdot$
$VP \rightarrow V\ NP \cdot NP$
$NP \rightarrow NP \cdot PP$
$NP \rightarrow \cdot Det\ N$
$NP \rightarrow \cdot Pron$
$NP \rightarrow \cdot NP\ PP$
$PP \rightarrow \cdot Prep\ NP$

*State 1*
$S \rightarrow NP \cdot VP$
$NP \rightarrow NP \cdot PP$
$VP \rightarrow \cdot V$
$VP \rightarrow \cdot V\ NP$
$VP \rightarrow \cdot V\ NP\ NP$
$VP \rightarrow \cdot VP\ PP$
$PP \rightarrow \cdot Prep\ NP$

*State 4*
$S' \rightarrow S \cdot$

*State 6*
$VP \rightarrow V \cdot$
$VP \rightarrow V \cdot NP$
$VP \rightarrow V \cdot NP\ NP$
$NP \rightarrow \cdot Det\ N$
$NP \rightarrow \cdot Pron$
$NP \rightarrow \cdot NP\ PP$

*State 9*
$VP \rightarrow VP\ PP \cdot$

*State 10*
$NP \rightarrow Det\ N \cdot$

*State 12*
$VP \rightarrow V\ NP\ NP \cdot$
$NP \rightarrow NP \cdot PP$
$PP \rightarrow \cdot Prep\ NP$

*State 13*
$PP \rightarrow Prep\ NP \cdot$
$NP \rightarrow NP \cdot PP$
$PP \rightarrow \cdot Prep\ NP$

Figure 2: The internal states of the toy grammar

are discussed in [5]. One of the main questions is that of equality or similarity between linguistic objects.

Consider constructing the non-kernel items using UG phrases following the dot in items already in the set for prediction. If such a phrase unifies with the LHS of a grammar rule and we add the new item with this instantiation, we need a mechanism to ensure termination — the risk is that we add more and more instantiated versions of the same item indefinitely. One might object that this is easily remedied by only adding items that are not subsumed by any previous ones. Unfortunately, this does not work, since it is quite possible to generate an infinite sequence of items none of which subsumes the other, see [9]. This problem can be solved by using so called "restrictors" to block out the feature propagation leading to non-termination, see [11], but still the number of items that are slight variants of one-another may be quite large. In her paper [5], Nakazawa proposes a simple and elegant solution to this problem:

> "While the CLOSURE procedure makes top-down predictions in the same way as before [using the full constraints of the unification grammar], new items are added without instantiation. Since only original productions in a grammar appear as items, productions are added as new items only once and the nontermination problem does not occur, as is the case of the LR parsing algorithm with atomic categories."

Unfortunately, even with this simplification, computing the non-kernel closure is quite time-consuming for large unification grammars.

Empty productions are a type of grammar rules that constitutes a notorious problem for parser developers. The LHS of these grammar rules have no realization in the input string since their RHS are empty. They are used to model movement as in the sentence *What$_i$ does John seek $e_i$?*, which is viewed as a transformation of *John seeks what?*. This is an example of left movement, since the word "what" has been moved to the left. Examples of right movement are rare in English, but frequent in other languages, the prime example being German subordinate clauses.

The particular unification grammar used keeps track of moved phrases by employing gap threading, i.e. by passing around a list of moved phrases to ensure that an empty production is only applicable if there is a moved phrase elsewhere in the sentence to license its use, see [6] pp. 125–129. As LR parsing is a parsing strategy employing bottom-up rule prediction, it is necessary to limit the applicability of these empty productions by the use of top-down filtering.

## 4 PARSER DESIGN

The parser was implemented and tested in SICStus Prolog using a version of the SRI Core Language Engine (CLE) [2] adapted to the air-travel information-service (ATIS) domain for a spoken-language translation task [8]. The CLE ordinarily employs a shift-reduce parser where each rule is tried in turn, although filtering using precompiled parsing tables makes it acceptably fast. The ATIS domain is a common ARPA testbench, and the CLE performance on it is comparable to that of other systems.

In fact, two slightly different versions of the parser were constructed, one for the original grammar, employing a mechanism for gap handling, as described in Section 4.2, and one for the learned grammar, where no such mechanism is needed, since this grammar lacks empty productions. Experiments were carried out over corpora of 100–200 test sentences, using SLR parsing tables, to measure the impact on parser performance of the various modifications described below.

A depth-first, backtracking LR parser was used were the parsing is split into three phases:

1. Phase one is the LR parsing phase. The grammar used here is the generalized unification grammar described in Section 4.1 below. The output is a parse tree indicating how the rules were applied to the input word string and what constraints were associated with each word.

2. Phase two applies the full constraints of the syntactic rules of the unification grammar and lexicon to the output parse tree of phase one.

3. Phase three applies the constraints of the compositional semantic rules of the grammar.

For the learned grammar, phase two and three coincide, since the learned rules include compositional semantic constraints. Each rule referred to in the output parse tree of phase one may be a generalization over several different rules of the unification grammar. Likewise, the constraints associated with each word can be a generalization over several distinct lexicon entries. In phase two, these different ways of applying the full constraints of the syntactic rules and the lexicon, and with the learned grammar also the compositional semantic constraints, are attempted non-deterministically.

The lookahead symbols, on the other hand, are ground Prolog terms. Firstly, this means that they can be computed efficiently in the LALR case. Secondly, this avoids trivial reduction ambiguities where a particular reduction is performed once for each possible mapping of the next word to a lookahead symbol. This is done by producing the set of all possible lookahead symbols for the next word at once, rather than producing one at a time non-deterministically. Each reduction is associated with another set of lookahead symbols. The intersection is taken, and the result is passed on to the next parsing cycle.

Prefix merging means that rules starting with similar phrases are processed together until they branch away. The problem with this in conjunction with a unification grammar is that it is not clear what "similar phrase" means. The choice made here is to regard phrases that map to the same CF symbol as similar:

**Definition**: *Two phrases are similar if they map to the same context-free symbol.*

Since the processing is performed by applying constraints incrementally and monotonically, where constraints are realized as Prolog terms and these are instantiated stepwise, it is important that a UG phrase map to the same CF symbol regardless of its degree of instantiation for this definition to be useful. The mapping of UG phrases to CF symbols used in the experiments was the naive one, where UG phrases mapped to their syntactic categories, (i.e. Prolog terms mapped to their functors), save that verbs with different complements (intransitive, transitive, etc.) were distinguished.

### 4.1 Generalization

The grammar used in phase one is not a context-free backbone grammar, nor the original unification grammar. Instead a generalized unification grammar is employed. This generalization is accomplish using anti-unification. This is the dual of unification — it constructs the least general term that subsumes two given terms — and was first described in [7]. This operation is often referred to as generalization in the computational-linguistics literature. If $T$ is the anti-unification of $T_1$ and $T_2$, then $T$ subsumes $T_1$ and $T$ subsumes $T_2$, and if any other term $T'$ subsumes both of $T_1$ and $T_2$, then $T'$ subsumes $T$. Anti-unification is a built-in predicate of SICStus Prolog and quite acceptably fast.

For each context-free rule, a generalized UG rule is constructed that is the generalization over all UG rules that map to that context-free rule. If there is only one such original UG rule, the full constraints of the unification grammar are applied already in phase one.

Similarly, the symbols of the action and goto tables are not context-free symbols. They are the generalizations of all relevant similar UG phrases. For example, each entry in the goto table will have as a symbol the generalization of a set of UG phrases. These UG phrases are those that map to the same context-free symbol; occur in a UG rule that corresponds to an item where this CF symbol immediately follows the dot; and in such a UG rule occur at the position immediately following the dot. For example, the symbol of the goto (or shift) entry for verbs between State 1 and State 6 of Fig. 2 is the anti-unification of the RHS verbs of the UG rules mapping to Rules 2, 3 and 4, e.g.

```
vp:[agr=Agr] => [v:[agr=Agr,sub=intran]].
vp:[agr=Agr] => [v:[agr=Agr,sub=tran],np:[agr=_]].
vp:[agr=Agr] =>
    [v:[agr=Agr,sub=ditran],np:[agr=_],np:[agr=_]].
```

which is `v:[agr=_,sub=_]`. Here the value of the subcategorization feature `sub` is left unspecified.

Lexical ambiguity in the input sentence is handled in the same way. For each word, a generalized phrase is constructed from all similar phrases it can be analyzed as. Again, if there is no lexical ambiguity within the CF symbol, the full UG constraints are applied. Nothing is done about lexical ambiguities outside of the same CF symbol, though.

In the experiments, using the UG constraints, instead of their generalizations, for the LR-parsing phase led to an increase in median normalized parsing time[1] from 3.1 to 3.8, i.e. by 20 %. This was also typically the case for the individual parsing times. In the machine-learning experiments, where normally several UG rules mapped to the same CF rule, this effect was more marked; it led to an increase in parsing time by a factor of five.

On the other hand, using truly context-free symbols for LR parsing actually leads to non-termination due to the empty productions. Even when banning empty productions, the parsing times increase by orders of magnitude; the vast majority (86 %) of the test sentences were timed out after ten minutes and still the normalized parsing time exceeded 100 in more than half (54 %) of the cases. This should be compared with the 0.220 figure using generalized UG constraints. In the machine-learning experiments, this lead to an increase in processing time by a factor 100.

### 4.2 Gap handling

A technique for limiting the applicability of empty productions is employed in the version for the original grammar. It is only correct for left movement. Since there are no empty productions in the learned grammar, there is no need for gap handling here.

The idea is that in order for an empty production to be applicable, some grammar rule must have placed a phrase corresponding to the moved one on the gap list. Thus a gap list is maintained where phrases corresponding to potential left movement are added whenever a state is visited where there is a "gap-adding phrase" immediately following the dot in any item. The elements of the gap list are the corresponding CF symbols. At this point the stack is "back-checked", as defined below, to see if the gap-adding rule really is applicable.

Back-checking means matching the prefixes of the kernel items against the stack in each state. The rationale for this is twofold. Firstly, capturing constraints on phrases previously obscured by grammar rules that have now branched off. Secondly, capturing feature agreement between phrases in prefixes of greater length than one. In general this was not useful; it simply resulted in a small overhead. In conjunction with gap handling, however, it proved essential.

The gap list is emptied after applying an empty production. This is not correct if several phrases are moved using the same gap list, or for conjunctions where the gap threading is shared between the conjuncts. For the former reason two different gap lists are employed — one for (auxiliary) verbs and one for maximal projections such as NPs, PPs, AdjPs and AdvPs.

In the experiments, omitting the gap-handling procedure led to non-termination; even just omitting the back-checking did so. By removing empty productions all together, the parsing times decreased an order of magnitude; the median normalized parsing time dropped to 0.220. This reduced the number of analyses of some sentences, and many sentences failed to parse at all. Nevertheless, this indicates that these rules have a strong adverse effect on parser performance.

## 5 COMPILER DESIGN

We turn now to the design of the compiler that constructs the parsing tables for the grammar. Although the compilation step involves a fair amount of pre- and postprocessing, the latter two consist of rather uninteresting menial tasks.

The parsing tables are constructed using the context-free backbone grammar, but also here there is opportunity for interleaving with the full UG constraints. The closure operation w.r.t. the non-kernel items is characteristic for the method.

The first point is viewing the closure operation as operating on sets. Consider the `closure/3` predicate of Fig. 3. [2] From an item already in the set, a set of non-kernel items is generated and its union with the original set is taken. The truly new items are added to the agenda driving the process.

The second point is matching the corresponding phrases of the unification grammar when predicting non-kernel items. This is done by the call to the predicate `check_ug_rules/4` of Fig. 3, and ensures that the

---

[1] The parsing time for the LR parser divided by the parsing time for the original parser.

[2] I am indebted to Mats Carlsson for this scheme. An efficient implementation of the primitive set operations such as union and intersection is provided by the ordered-set-manipulation package of the SICStus library. These primitives presuppose that the sets are represented as ordered lists and consist of ground terms.

```
closure(Set,Closure) :-
   closure(Set,Set,Closure).

closure([], Closure, Closure).
closure([Item|Items], Set0, Closure) :-
   findall(NkItem,
           n_k_item(Item,NkItem),
           NkItems),
   union(Set0,NkItems,Set1,NewItems),
   merge(NewItems,Items,Items1),
   closure(Items1,Set1,Closure).

n_k_item(item(Rule1,_,RHS0,RHS),
         item(Rule2,LHS2,RHS2,RHS2)) :-
   RHS = [LHS2|_],
   cf_rule(Rule2,LHS2,RHS2),
   check_ug_rules(Rule1,Rule2,RHS0,RHS).
```

Figure 3: The non-kernel closure function

phrase immediately following the "dot" in some UG rule mapping to Rule1 unifies with the LHS of some UG rule mapping to Rule2. In item(Rule,LHS,RHS0,RHS), Rule is an atomic rule identifier and RHS0 and RHS form a difference list marking the position of the dot.

This is a compromise between performing the closure operation with full UG constraints and performing it efficiently, and achieves the same net effect as the method in Section 3 advocated by Nakazawa. Especially in the machine-learning application, where rather large grammars are used, compiler performance is a most critical issue.

In the experiments, omitting the checking of UG rules when performing the closure operation leads to non-termination when parsing. This is because the back-checking table for the gap handler becomes too general. For the learned grammar, this made constructing the internal states prohibitively time-consuming.

## 6 SUMMARY

The design of the LR parser and compiler is based on interleaving context-free processing with applying the full constraints of the unification grammar.

Using a context-free description-level has the advantages of providing a criterion for similarity between UG phrases, allowing efficient processing both at compile time and runtime, and providing a basis for probabilistic analysis. The former makes prefix merging, which is the very core of LR parsing, well-defined for unification grammars, and enables using a generalized unification grammar in the LR parsing phase, which is one of the major innovations of the scheme. This and prefix merging are vital when working with the learned grammar since many rules overlap totally or partially on the context-free level.

Interleaving context-free processing with applying the full constraints of the unification grammar to prune the search space restores some of the predictive power lost using a context-free backbone grammar. In particular, using the full UG constraints "inside" the non-kernel closure operation to achieve the effect of using the unification grammar itself for performing this operation constitutes another important innovation.

The experiments emphasize the importance of restricting the applicability of empty productions through the use of top-down filtering. Thus the main remaining issue is to improve the gap handling mechanism to perform real gap threading.

## ACKNOWLEDGEMENTS

I wish to thank Mats Carlsson for valuable advice on Prolog implementation issues and Ivan Bretan, Robert Moore and Manny Rayner for clear-sighted comments on draft versions of this article and related publications, and for useful suggestions to improvements.